\documentclass[10pt, a4paper]{article}

\usepackage[final]{lrec2026} 

\usepackage{tikz}
\usepackage{multirow}
\usepackage{booktabs}
\usepackage{amsmath}

\usetikzlibrary{shapes, arrows.meta, positioning, fit}

\title{SENS-ASR: Semantic Embedding injection in Neural-transducer for Streaming Automatic Speech Recognition}

\name{Youness Dkhissi$^{1,2}$, Valentin Vielzeuf$^{1}$, Elys Allesiardo$^{1}$, Anthony Larcher$^{2}$} 

\address{1. Orange Innovation, 4 Rue du Clos Courtel, 35510 Cesson-Sévigné, France\\
        2. LIUM, Le Mans Université Avenue Olivier Messiaen,72085 Le Mans CEDEX 9, France \\
         \{youness.dkhissi, valentin.vielzeuf, elys.allesiardo\}@orange.com, anthony.larcher@univ-lemans.fr\\}

\abstract{
Many Automatic Speech Recognition (ASR) applications require streaming processing of the audio data. In streaming mode, ASR systems need to start transcribing the input stream before it is complete, i.e., the systems have to process a stream of inputs with a limited (or no) future context. Compared to offline mode, this reduction of the future context degrades the performance of Streaming-ASR systems, especially while working with low-latency constraint.
In this work, we present \textbf{SENS-ASR}, an approach to enhance the transcription quality of Streaming-ASR by reinforcing the acoustic information with semantic information.
This semantic information is extracted from the available past \textit{frame-embeddings} by a context module. This module is trained using knowledge distillation from a sentence embedding Language Model fine-tuned on the training dataset transcriptions. Experiments on standard datasets show that SENS-ASR significantly improves the Word Error Rate on small-chunk streaming scenarios.
 \\ \newline \Keywords{streaming automatic speech recognition, semantic representation, neural transducer, large language model.} }

\begin{document}

\maketitleabstract

\section{Introduction}

In recent years, End-to-End (E2E) ASR approaches such as Connectionist Temporal Classification (CTC) \cite{graves2006connectionist}, Attention Encoder-Decoder (AED) \cite{chan2016listen,radford2023robust} and Recurrent Neural Network Transducer (RNN-T) \cite{graves2012sequence} have gained a lot of popularity compared to hybrid models \cite{bourlard2012connectionist}, especially due to the emergence of transformer-based architectures \cite{vaswani2017attention}. These approaches give great results in terms of transcription quality when having access to the full speech audio as input. However, in Streaming-ASR \cite{variani2022global} where models should begin transcribing without having the full speech context, the deployment of E2E models presents challenges.
Most of the proposed approaches suffer from severe performance degradation as they use causal masking to learn transcribing without relying on the future context \cite{moritz21_interspeech,wang2024sscformer}.

Among the works that try to tackle the problem of transcription quality degradation in Streaming-ASR,  \cite{yu2020dual} proposed a streaming-offline process to teach a unique model to transcribe with and without having the full speech context. When operating in streaming mode, this approach enhances the frame representations generated by the encoder with information gathered from the model's offline training, which utilizes the full context. 

Other works change the causal attention mask into a chunk-wise attention mask that enables the model to benefit from a full attention operation on well-defined chunks of frames, rather than being limited to causal attention \cite{gulzar2023ministreamer,wang2024sscformer}. This masking technique improves the predictions of the model, especially for the frames located at the beginning of each chunk. Nevertheless, this approach does not compensate for the lack of future context when processing the last frames of each chunk. 
This drawback motivated the work of  \cite{zeineldeen2024chunked} in which a chunk-wise attention mechanism with lookahead enables the model to take into account some frames of the future chunk (i.e., part of the future context) when encoding the frames from the current chunk. However, this adaptation increases the latency induced by waiting for extra frames as well as the computational cost due to the redundancy of encoding frames used in adjacent chunks. 
These drawbacks encourage other works to generate a simulated lookahead based on past frames rather than using a true one \cite{an2022cuside,zhao2024cuside}.

All of these previous works try to improve Streaming-ASR models based on acoustic features only. In fact,  \cite{choi2024self} and  \cite{sanabria2023measuring} show that the embeddings generated for the audio frames mostly embed acoustic information rather than semantic information. In addition,  \cite{kim2021improved} indicates that RNN-T, in particular, has poor results in modeling long-range linguistic information. For this reason, the existing approaches use rescoring methods based on External Language Models during inference to address the lack of semantic information in the embeddings generated by the encoder and thus improve the transcription quality.

Other recent works try to exploit Large Language Models for the ASR task \cite{chen2024salm} and also in streaming context \cite{tsunoo2024decoder}.
However, using some of these LLMs in the core of ASR architecture brings many doubts about their real effectiveness, as they are evaluated on public test datasets whose transcriptions could potentially be used in the training of these LLMs. These leaks are found and confirmed on many Natural Language Processing datasets \cite{balloccu2024leak, xu2024benchmark}, which increases the possibility that transcriptions of public ASR datasets have been used in the training of some LLMs. A recent work \cite{tseng2025evaluation} shows that a substantial amount of the LibriSpeech \citelanguageresource{panayotov2015librispeech} and Common Voice \citelanguageresource{ardila-etal-2020-common} evaluation sets, which are amongst the most used datasets for speech recognition, appear in public LLM pre-training corpora. This specifically questions the results and the improvements obtained by LLM based models on these datasets.

We propose \textbf{SENS-ASR}, a novel framework that directly addresses the semantic deficiency of Streaming-ASR by injecting semantic information into the encoder’s \textit{frame-embeddings}. Unlike prior approaches that treat acoustic and linguistic modeling as separate or loosely coupled components, SENS-ASR introduces a dedicated context module that operates in real time, generating semantic embeddings from the history of past acoustic frames. This module is trained via knowledge distillation from a sentence embedding language model, which is itself fine-tuned on the target ASR domain to ensure relevance and robustness.

The core idea behind SENS-ASR is to bridge the gap between local acoustic features and global semantic context, enabling the encoder to produce frame representations that are both acoustically and semantically informative. By enriching each frame with semantic context derived from the preceding audio, SENS-ASR empowers the decoder to make more accurate and coherent predictions, even under strict streaming constraints.

Overall, our paper comes with the following contributions: 
\textbf{(a)} a transducer model equipped with an additional context module, designed to inject semantic information into the representation of the frames and \textbf{(b)} a finetuning protocol of the sentence embedding allowing us to then train the context module.
 
\section{Proposed method}
\begin{figure}
    \centering
    \resizebox{0.5\textwidth}{!}{\begin{tikzpicture}[node distance=2cm, auto, thick, 
    box/.style={draw, rectangle, rounded corners, minimum height=1cm, minimum width=2cm, align=center},
    arrow/.style={-Stealth, thick},
    dashedbox/.style={draw, dashed, circle, align=center},
    label/.style={font=\footnotesize, align=center}]

    \node (lm) [box] {Sentence Embedding \\ Model};
    \node (semantic_emb) [above=2cm of lm] {Semantic Embedding};
    \node (distillation) [right=0.5cm of semantic_emb, dashedbox, label=above:Distillation] {Distillation\\Loss\\$\alpha\mathcal{L}_{MSE}$};
    \node (context_emb) [right=0.5cm of distillation] {Context Embedding $C$};
    \node (context) [below=2cm of context_emb, box, fill=green!20] {Context Module};
    \node (enc_emb) [below=1cm of context] {Frame-Embeddings $H$};
    \node (encoder) [below=2cm of context, box] {Encoder};
    \node (joint) [right=2cm of context, box] {Joint};
    \node (predictor) [below=2cm of joint, box] {Predictor\\ (RNN)};
    \node (prev_pred) [below=1cm of predictor] {Previous Prediction};
    \node (token) [above=2cm of joint] {\textbf{Prediction}};
    \node (asr_loss) [above=0.5cm of token, dashedbox, label=above:Distillation] {ASR Loss\\$\mathcal{L}_{RNN-T}$};
    \node (frames) [below=1cm of encoder] {Frames $X$};
    \node (spoken_tokens) [below=2cm of lm, text width=3cm, align=center] {Full audio transcription};
    \node (training_phase) [below=0.5cm of spoken_tokens, text width=2.5cm, align=center, red] {During training stage only};

    \draw[arrow] (lm) -- (semantic_emb);
    \draw[arrow] (semantic_emb) -- (distillation);
    \draw[arrow] (context_emb) -- (distillation);
    \draw[arrow] (context) -- (context_emb);
    \draw[arrow] (encoder) -- (enc_emb);
    \draw[arrow] (enc_emb) -- (context);
    \draw[arrow] (context_emb) -- (joint);
    \draw[arrow] (enc_emb) -- (joint);
    \draw[arrow] (joint) -- (token);
    \draw[arrow] (token) -- (asr_loss);
    \draw[arrow] (predictor) -- (joint);
    \draw[arrow] (frames) -- (encoder);
    \draw[arrow] (prev_pred) -- (predictor);
    \draw[arrow] (spoken_tokens) -- (lm);

    \node[draw=red, dashed, fit=(lm) (spoken_tokens) (semantic_emb) (distillation), inner sep=0.5cm, label=above left:Training Phase] {};

\end{tikzpicture}}
    \caption{Architecture of the SENS-ASR system using an RNN-T model and a context module.
    Components in the red dashed-rectangle are only used during training. Components in dashed-circles are the parts of the system global loss.}
    \label{fig:semantic_context_architecture}
    \vspace{-0.2cm}
\end{figure}
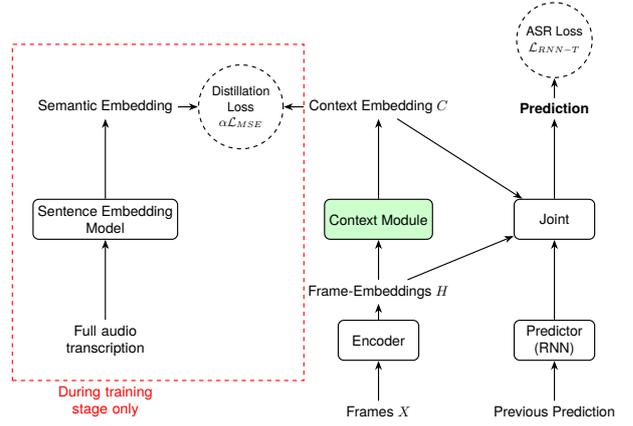

Figure \ref{fig:semantic_context_architecture} shows the training process and the new components applied to an RNN-T model. \textbf{SENS-ASR} operates in two stages: at training stage, a \textit{context module} learns to model the semantic context by distillation from a teacher Sentence Embedding Model; at inference stage, this module extracts semantic information to enrich the acoustic \textit{frame-embeddings}.

\subsection{Semantic Context}
Let $X = \{x^{(\gamma)}_i\}$ be the sequence of frames we input per chunk into the ASR encoder, with $x^{(\gamma)}_i$ being the frame $i$ in the chunk $\gamma$. 
All chunks have a fixed length $s$.
The ASR encoder generates a \textit{frame-embedding} for each frame, $x^{(\gamma)}_i$, using a chunk-wise attention. In the decoding step, \textbf{SENS-ASR} injects, in each \textit{frame-embedding}, a semantic information extracted from the past-context of this frame.
Each \textit{frame-embedding}, $h^{(\gamma)}_i$, is computed as a function of all frames from past chunks $0$ to $\gamma - 1$
plus all frames from the current chunk $\gamma$.
%
\begin{equation}
\begin{cases}
    H = Encoder(X) \\
    h^{(\gamma)}_i = f(x^{(0)}_0, \dots, x^{(\gamma)}_{s-1})\\
\end{cases}
\end{equation}

To inject semantic information into the decoding process, 
past \textit{frame-embeddings} are processed by the context-module
to generate one context-embedding, $C^{(\gamma)}_i$, for each $h^{(\gamma)}_i$.
To reduce the computational complexity, a unique context embedding $C^{(\gamma)}$ 
is computed for each chunk, $\gamma$, using the past $P$ chunks.
\begin{equation}
    C^{(\gamma)} = Context\_Module(x^{(\gamma - P)}_0, \dots, x^{(\gamma - 1)}_{s-1})
\end{equation}

This module is trained to have an output similar to that of the Sentence Embedding Model. 
This model takes the transcription of the full audio and generates a sentence embedding of this transcription. 

This context-embedding can be computed by techniques such as max, average or attention pooling.
Based on  \cite{chen-etal-2018-enhancing}, we use attention pooling in our semantic context module. 
More precisely, we apply successive cross-attention operations on \textit{frame-embeddings} to produce a single vector.

After computing the semantic context embedding for a chunk, $\gamma$, 
we concatenate it with each \textit{frame-embedding}, $h^{(\gamma)}_i$, of this chunk before passing them to the joint network together with predictor network output to generate the prediction as described in  \cite{graves2012sequence}.

We train our model using the following loss:
\begin{equation}
    \mathcal{L}_{SENS-ASR}=\mathcal{L}_{RNN-T}+\alpha \mathcal{L}_{MSE}
\end{equation}

where $\mathcal{L}_{RNN-T}$ is the standard transducer loss for the ASR task, $\alpha$ is a scalar and $\mathcal{L}_{MSE}$ is the Mean Square Error loss used to train the context module to mimic the semantic embedding generated by the Teacher Sentence Embedding Model.

\subsection{Teacher Sentence Embedding Model fine-tuning}
\label{sec:lm_fine-tuning}
To improve the quality of the semantic information extracted in our application domain, it is necessary to fine-tune the teacher Sentence Embedding Model that will guide the training of our context module. This involves generating pairs of sentences by paraphrasing the transcriptions from the training dataset, which will be used to fine-tune the Language Model with the objective of reinforcing the similarity between embeddings of these pairs. To avoid neural collapse \cite{papyan2020prevalence}, we design a dataset with positive and negative pairs of sentences.

\subsubsection{Paraphrasing protocol}
To perform fine-tuning, we create a set of text pairs (sentence A, sentence B) where sentence A is the transcription of an audio segment, while sentence B is a paraphrase artificially generated. 
Paraphrases can be generated by lexical replacement, back-translation \cite{liu2020survey} or by using Language Models 
(small ones fine-tuned for paraphrase generation or even LLMs). 
We choose to use LLM paraphrasing as other techniques generate text which is not far enough from the original in terms of information order and vocabulary.

\begin{figure}
    \centering
    \resizebox{\linewidth}{!}{
    \includegraphics[width=\linewidth]{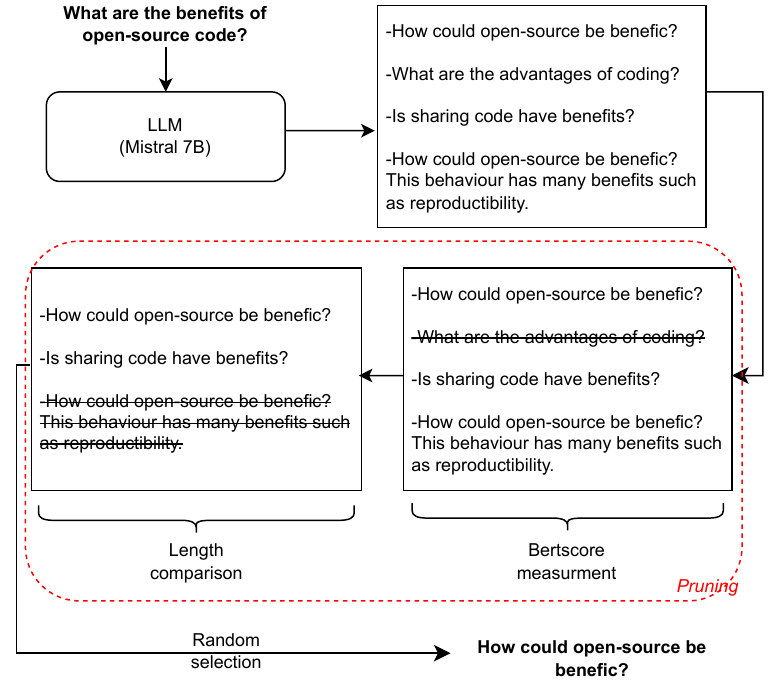}}
    \caption{Example of the proposed LLM paraphrasing process. The text in \textbf{bold} refers to the input and the output, while the strikethrough text represents the rejected paraphrases in each pruning step.}
    \label{fig:llm_paraphrasing}
    \vspace{-0.2cm}
\end{figure}
However, LLMs could hallucinate, i.e., 
rewrite the original text in addition to the proposed paraphrase or respond to the content of the original text if it is truncated or in a question form. 
To avoid these hallucinations, as described in Figure \ref{fig:llm_paraphrasing},
we generate multiple paraphrases and filter them according to two criteria: i) we discard sentences whose BERTScore with the original text is less than 0.5 \cite{zhang2019bertscore}; ii) paraphrases that are too long with respect to the original text (2x longer or more) are also discarded.
In the end, the paraphrase is randomly chosen among the remaining sentences.

We could not control these hallucinations while paraphrasing truncated sentences. That is why we choose to apply the fine-tuning of the Teacher Sentence Embedding Model using full audio transcription, which is more challenging for the context module to mimic. Fortunately, this choice is not an obstacle to have an efficient training as shown in Section \ref{sec:results_dicussion}.

\subsubsection{Avoiding neural collapse during fine-tuning}
 
In the previous section, we built positive pairs of sentences that share similar semantic content.
We now create negative pairs by associating an original text transcribed from an audio utterance with the transcription of another audio utterance or its paraphrase.

We create triplets (sentence A, sentence B, \textit{label}) where \textit{label} is a Cosine Similarity objective
between sentence A and sentence B which is randomly chosen as follows:
\begin{itemize}
    \item $label\sim \mathcal{U}(0.8, 1)$ if (sentence A, sentence B) is a positive pair.
    \item $label\sim \mathcal{U}(-0.2, 0.2)$ if (sentence A, sentence B) is a negative pair.
\end{itemize}
In our training corpus, each speaker tends to talk about a limited number of topics.
Thus, associating transcriptions from a single speaker in a negative pair would induce a bias.
We constrain our negative pairs to come from different speakers.

Positive pairs represent $\frac{2}{3}$ of the fine-tuning dataset, while the remaining $\frac{1}{3}$ represents the portion of negative pairs. These proportions are chosen to optimize the model training while avoiding neural collapse.

\subsection{Dynamic Chunk Training}

Dynamic Chunk Training (DCT) \cite{zhang2020unified} is a technique designed to enable an Automatic Speech Recognition (ASR) model to operate effectively in both streaming and offline modes by exposing it to a variety of context lengths during training. This approach relies on constructing a mask matrix \( M \in \{0,1\}^{T \times T} \), where \( T \) is the total number of frames in the input sequence. Each element \( m_{t,u} \) of the mask indicates whether, at time step \( t \), the model is permitted to attend to the input at time \( u \).

During training, a chunk size \( S \) is sampled randomly from the interval \([1, T]\). To control the amount of past context the model can attend to, the left context is limited to \( P \) chunks, with \( P \) sampled from the set \(\{0, \lceil T / S \rceil\}\). The mask matrix \( M \) is then constructed such that:

\begin{equation}
m_{t,u} = 
\begin{cases}
1, & \text{if } \left\lfloor \frac{t}{S} \right\rfloor - P \leq \left\lfloor \frac{u}{S} \right\rfloor \leq \left\lfloor \frac{t}{S} \right\rfloor, \\
0, & \text{otherwise}.
\end{cases}
\end{equation}

Where $\left\lfloor \frac{t}{S} \right\rfloor$ is the chunk index of $t$ and $\left\lfloor \frac{u}{S} \right\rfloor$ is the chunk index of $u$.

This formulation ensures that, at each time step \( t \), the model's attention is confined to the current chunk and a limited number of previous chunks, effectively controlling the computational complexity of the attention mechanism. When \( S = T \), the mask reduces to all ones, allowing the model to attend to the entire sequence, which corresponds to offline processing. Conversely, smaller values of \( S \) simulate streaming conditions with limited context, enabling the model to generalize across both streaming and non-streaming scenarios.

In practice, this dynamic chunking strategy is integrated into the training process by randomly selecting \( S \) and \( P \) for each batch, thereby exposing the model to a diverse range of context lengths. This approach facilitates the development of models capable of flexible inference, maintaining high accuracy while reducing computational costs during decoding.

\section{Experiments}
\begin{table*}[!htbp]
  \caption{the Word Error Rate (WER) measurements on the LibriSpeech test-clean, test-other and TEDLIUM-2 datasets with different inference chunk sizes using a unique model trained on Dynamic Chunk Training (DCT) interval of [160ms;1,280ms].The values in brackets correspond to the difference compared to the baseline and the values in square brackets represent the confidence interval of the result}
  \label{tab:baseline_vs_semantic_context}
  \centering
\begin{tabular}{|c|c|ccccc|}
\cline{1-7}
\multirow{2}{*}{Dataset} &
  \multirow{2}{*}{Model} &
  \multicolumn{5}{c|}{Chunk size}  \\ 
 &
   &
  160ms &
  320ms &
  640ms &
  1,280ms &
  \multicolumn{1}{c|}{Full-context} \\ \cline{1-7}
\multirow{2}{*}{\begin{tabular}[c]{@{}c@{}}LibriSpeech \\test-clean\end{tabular}} &
  Baseline &
  \multicolumn{1}{c|}{\begin{tabular}[c]{@{}c@{}}7.55\\ \footnotesize{{[}7.24;7.87{]}}\end{tabular}} &
  \multicolumn{1}{c|}{\begin{tabular}[c]{@{}c@{}}4.82\\ \footnotesize{{[}4.57;5.06{]}}\end{tabular}} &
  \multicolumn{1}{c|}{\begin{tabular}[c]{@{}c@{}}3.90\\ \footnotesize{{[}3.66;4.12{]}}\end{tabular}} &
  \multicolumn{1}{c|}{\begin{tabular}[c]{@{}c@{}}3.49\\ \footnotesize{{[}3.27;3.69{]}}\end{tabular}} &
  \begin{tabular}[c]{@{}c@{}}2.90\\ \footnotesize{{[}2.71;3.11{]}}\end{tabular} \\ \cline{2-7} 
 &
  SENS-ASR &
  \multicolumn{1}{c|}{\begin{tabular}[c]{@{}c@{}}7.21\textbf{(-0.34)}\\ \footnotesize{{[}6.89;7.53{]}}\end{tabular}} &
  \multicolumn{1}{c|}{\begin{tabular}[c]{@{}c@{}}4.73\textbf{(-0.09)}\\ \footnotesize{{[}4.48;4.99{]}}\end{tabular}} &
  \multicolumn{1}{c|}{\begin{tabular}[c]{@{}c@{}}3.83\textbf{(-0.07)}\\ \footnotesize{{[}3.61;4.03{]}}\end{tabular}} &
  \multicolumn{1}{c|}{\begin{tabular}[c]{@{}c@{}}3.44\textbf{(-0.05)}\\ \footnotesize{{[}3.24;3.65{]}}\end{tabular}} &
  \begin{tabular}[c]{@{}c@{}}2.93(+0.03)\\ \footnotesize{{[}2.73;3.14{]}}\end{tabular} \\ \cline{1-7}

\multirow{2}{*}{\begin{tabular}[c]{@{}c@{}}LibriSpeech \\test-other\end{tabular}} &
  Baseline &
  \multicolumn{1}{c|}{\begin{tabular}[c]{@{}c@{}}18.34\\ \footnotesize{{[}17.80;18.88{]}}\end{tabular}} &
  \multicolumn{1}{c|}{\begin{tabular}[c]{@{}c@{}}12.41\\ \footnotesize{{[}11.98;12.87{]}}\end{tabular}} &
  \multicolumn{1}{c|}{\begin{tabular}[c]{@{}c@{}}9.70\\ \footnotesize{{[}9.32;10.10{]}}\end{tabular}} &
  \multicolumn{1}{c|}{\begin{tabular}[c]{@{}c@{}}8.39\\ \footnotesize{{[}8.03;8.78{]}}\end{tabular}} &
  \begin{tabular}[c]{@{}c@{}}6.76\\ \footnotesize{{[}6.44;7.07{]}}\end{tabular} \\ \cline{2-7} 
 &
  SENS-ASR &
  \multicolumn{1}{c|}{\begin{tabular}[c]{@{}c@{}}17.89\textbf{(-0.45)}\\ \footnotesize{{[}17.38;18.47{]}}\end{tabular}} &
  \multicolumn{1}{c|}{\begin{tabular}[c]{@{}c@{}}12.11\textbf{(-0.30)}\\ \footnotesize{{[}11.69;12.60{]}}\end{tabular}} &
  \multicolumn{1}{c|}{\begin{tabular}[c]{@{}c@{}}9.66\textbf{(-0.04)}\\ \footnotesize{{[}9.30;10.07{]}}\end{tabular}} &
  \multicolumn{1}{c|}{\begin{tabular}[c]{@{}c@{}}8.55(+0.16)\\ \footnotesize{{[}8.21;8.94{]}}\end{tabular}} &
  \begin{tabular}[c]{@{}c@{}}6.90(+0.14)\\ \footnotesize{{[}6.56;7.22{]}}\end{tabular} \\ \cline{1-7}

\multirow{2}{*}{TEDLIUM-2} &
  Baseline &
  \multicolumn{1}{c|}{\begin{tabular}[c]{@{}c@{}}16.52\\ \footnotesize{{[}15.83;17.14{]}}\end{tabular}} &
  \multicolumn{1}{c|}{\begin{tabular}[c]{@{}c@{}}11.94\\ \footnotesize{{[}11.39;12.52{]}}\end{tabular}} &
  \multicolumn{1}{c|}{\begin{tabular}[c]{@{}c@{}}10.04\\ \footnotesize{{[}9.50;10.55{]}}\end{tabular}} &
  \multicolumn{1}{c|}{\begin{tabular}[c]{@{}c@{}}9.00\\ \footnotesize{{[}8.52;9.51{]}}\end{tabular}} &
  \begin{tabular}[c]{@{}c@{}}8.33\\ \footnotesize{{[}7.88;8.84{]}}\end{tabular} \\ \cline{2-7} 
 &
  SENS-ASR &
  \multicolumn{1}{c|}{\begin{tabular}[c]{@{}c@{}}15.60\textbf{(-0.92)}\\ \footnotesize{{[}14.98;16.23{]}}\end{tabular}} &
  \multicolumn{1}{c|}{\begin{tabular}[c]{@{}c@{}}11.82\textbf{(-0.12)}\\ \footnotesize{{[}11.25;12.35{]}}\end{tabular}} &
  \multicolumn{1}{c|}{\begin{tabular}[c]{@{}c@{}}9.79\textbf{(-0.25)}\\ \footnotesize{{[}9.28;10.30{]}}\end{tabular}} &
  \multicolumn{1}{c|}{\begin{tabular}[c]{@{}c@{}}8.96\textbf{(-0.04)}\\ \footnotesize{{[}8.49;9.43{]}}\end{tabular}} &
  \begin{tabular}[c]{@{}c@{}}8.33\\ \footnotesize{{[}7.85;8.85{]}}\end{tabular} \\ \cline{1-7}
\end{tabular}
\end{table*}

\subsection{Evaluation protocol}

We conduct our experiments on two widely used public datasets: LibriSpeech \cite{panayotov2015librispeech} that contains 960 hours of read English speech and TEDLIUM-2 \citelanguageresource{zhou2020rwth} that contains 207 hours of TED Talks in order to show the effectiveness of our approach on more spontaneous speech.

To evaluate our approach on the Streaming-ASR task, we use the Word Error Rate metric. Also, to show the significance of the results, each result is presented together with its confidence interval\footnote{https://github.com/luferrer/ConfidenceIntervals}. These intervals are calculated using the bootstrapping method with 1,000 bootstrap sets. They were also calculated between $2.5$ and $97.5$ percentiles to exclude outliers.

\subsection{Model configuration}

All experiments\footnote{https://github.com/Orange-OpenSource/sens-asr} were conducted using the SpeechBrain toolkit \cite{ravanelli2021speechbrain}. The baseline used for Streaming-ASR task and our proposed system are Recurrent Neural Network Transducer (RNN-T \cite{graves2012sequence}) which share the following components:
\begin{itemize}
    \item \textbf{2 Convolutional layers} with kernel size of 2 and stride of 2, which downsamples the frame rate by 4.
    \item \textbf{12-layer Conformer encoder} \cite{gulati2020conformer} having 512-dimensional input where each layer is composed of: feed-forward network of size 2048, convolution bloc having kernel size of 31 with stride of 1 and a self-attention bloc with 8 attention heads.
    \item \textbf{Predictor network} of 1-layer LSTM \cite{graves2012long} with hidden size of 512.
    \item \textbf{Joint network} where a projection of dimension 640 is applied to the encoder output and another to the predictor output. Then we sum the 2 resulting vectors.
\end{itemize}

 Our proposed SENS-ASR architecture includes an additional \textbf{Context module} composed of an attention pooling of a 3-layer transformer decoder followed by a linear projection of dimension 768 to match the output of the teacher Sentence Embedding Model.

During the training stage, the RNN-T loss with Fastemit regularization \cite{yu2021fastemit} ($\lambda=0.006$) is used to optimize the model latency. The distillation loss has a weight of $\alpha=0.2$. This value was chosen empirically 
to ensure the convergence of the distillation loss without affecting the convergence speed of $\mathcal{L}_{RNN-T}$.
In addition, an Adam optimizer is set with a learning rate of $0.0008$ and a weight decay of $0.01$.

Both baseline RNN-T and SENS-ASR systems are trained only once using Dynamic Chunk Training (DCT) \cite{zhang2020unified}.
During training, 60\% of the batches use a chunk size randomly chosen from the interval [160ms;1,280ms] while other batches contain the full speech context.

The context module is trained with MPnet model \cite{song2020mpnet} which has previously been pre-trained and fine-tuned on 1 billion pairs of sentences\footnote{https://huggingface.co/sentence-transformers/all-mpnet-base-v2}. 
We perform a second stage of fine-tuning, as explained in section \ref{sec:lm_fine-tuning}, using Mistral 7B Large Language Model to generate the paraphrases.

In the inference stage, we use \textbf{greedy search without an External Language Model rescoring} in all experiments to highlight the benefit of our contributions.

\begin{table*}[!htbp]
  \caption{Comparison between State Of The Art models and SENS-ASR in terms of WER on LibriSpeech test-clean. For SENS-ASR, the chunk sizes with (*) were not the only targeted sizes during training stage.}
  \label{tab:sota_wer_librispeech}
  \centering
  \begin{tabular}{|c|c|c|c|}
\hline
Model & \begin{tabular}[c]{@{}c@{}}Additional\\ characteristics\end{tabular} & \begin{tabular}[c]{@{}c@{}}Chunk\\size(ms)\end{tabular} & \begin{tabular}[c]{@{}c@{}}WER\\ (\%)\end{tabular} \\ \hline
trimtail  \cite{song2023trimtail} & -  & 640  & 4.68   \\ \hline
ZeroPrompt  \cite{song2023zeroprompt} & - & 640 & 4.41 \\ \hline
CA transformer  \cite{li2023cumulative}   & Beam size=10 & 1,280 & 3.8\\ \hline
Delay penalized transducer  \cite{kang2023delay} & $\lambda = 0.006$ & 640 & 3.74 \\ \hline
Streamable decoder-only  \cite{tsunoo2024decoder} & \begin{tabular}[c]{@{}c@{}}Beam size=10\\ w LM\end{tabular} & 1,600 & 3.2 \\ \hline \hline
\multirow{2}{*}{SENS-ASR(ours)} & $\lambda = 0.006$ & 640* & 3.83 \\ \cline{2-4}
 & $\lambda = 0.006$ & 1,280* & 3.44 \\ \hline
\end{tabular}
\end{table*}
\subsection{Results and discussions}
\label{sec:results_dicussion}
Table \ref{tab:baseline_vs_semantic_context} shows the performance of our SENS-ASR system compared to the baseline. 
Both systems are trained only once with DCT but used with a fixed chunk size at inference.
The results indicate that SENS-ASR significantly reduces WER when inference is conducted with small chunk sizes (160ms and 320 ms). Then, insignificant improvements are observed with larger chunk sizes (640ms and 1,280ms).Finally, there is no improvement using the full-context audio. 
This trend can be attributed to the fact that larger chunks often contain sufficient acoustic information for a complete word(s) pronunciation.  In particular, the WER for the SENS-ASR model is 7.21\% for a chunk size of 160ms, which is an absolute reduction of 0.34\% compared to the baseline, and 3.44\% for 1,280ms, showing an absolute decrease of 0.05\%. Also, for test-other, SENS-ASR reduces the WER by an absolute value of 0.45\% compared to the baseline for 160ms while it gets a slight increase of 0.16\% for 1,280ms. Similarly, for the TEDLIUM-2 dataset, SENS-ASR achieves a WER of 15.60\% at a chunk size of 160ms, which is an absolute reduction of 0.92\% compared to the baseline, and 8.96\% for 1,280ms, indicating a decrease of 0.04\%. These results demonstrate consistent improvements across datasets for small chunk sizes.

We compare in Table \ref{tab:sota_wer_librispeech} our model with the State Of The Art models in Streaming-ASR, tested on LibriSpeech test-clean. Overall, the proposed model achieves competitive ASR performance and sometimes even better than models that use larger chunk size. 
It is difficult to draw a definitive conclusion on which model is better due to the differences on architectures used and the configurations that are discussed or not, in the papers. Yet, our SENS-ASR model, which has been trained only once with DCT is competitive on all reported chunk sizes compared with models trained specifically for this size.

To further push the comparison, we train two SENS-ASR models, each on a unique chunk size\footnote{thus training 40\% of batches with full context and 60\% of unique chunk size} , 320 ms and 640 ms, later used in inference and compare them to the model we train using Dynamic Chunk Training (DCT).
While testing on Librispeech test-clean, the model improves in terms of WER when training on a targeted chunk size. The model trained on 160 ms chunk size has a WER of 4.58\%, i.e., an absolute WER reduction of \textbf{2.63\%}. Also, the model trained on 320 ms chunk size has an absolute WER reduction of \textbf{0.55\%} compared to the model trained with DCT. 

Although the targeted chunk size training gives a big improvement in terms of transcription quality, this technique is \textbf{more computationally costly} (one training for each chunk size). Moreover, it \textbf{reduces the model robustness} while testing it with unseen chunk sizes during training.

\subsection{Error analysis}
To complete our analysis, we propose an error analysis. We compare the baseline and SENS-ASR with the 160 ms chunk size setup. We aim to better identify the potential benefits and drawbacks of the proposed method compared to a basic approach.

\begin{table}[!htbp]
  \caption{Error comparison between SENS-ASR and the baseline using the chunk size of 160 ms by using WER and by each type of error. The values in brackets written in \textbf{bold} correspond to the relative difference compared to the baseline.}
  \label{tab:comparision_type_error}
  \centering
  \resizebox{\linewidth}{!}{
  \begin{tabular}{c|c|c}
    &\textbf{Baseline} &
                        \textbf{SENS-ASR}\\
    \midrule
    \textbf{WER (\%)} & 7.55  & 7.21              \\
    \midrule
    \textbf{Number of Insertions} & 507  & 403 \textbf{(-20.51\%)}              \\
    \textbf{Number of Deletions} & 374  & 370 \textbf{(-1.07\%)}              \\
    \textbf{Number of Substitutions} & 3091  & 3020 \textbf{(-2.30\%)}              \\
    \bottomrule
  \end{tabular}}
\end{table}

We compare the errors of models by type of edition, as we show in the Table \ref{tab:comparision_type_error}. We may consider that our approach has succeeded in reducing a significant number of insertions (compared to the baseline). This leads us to the idea that adding a semantic embedding helps reduce the tendency for overly verbose transcriptions generated by the baseline.
\section{Conclusion}
In this paper, we present SENS-ASR, a method that enhances streaming Automatic Speech Recognition (ASR) by integrating semantic information into \textit{frame-embeddings}. Our approach addresses the limitations of traditional models, resulting in noticeable improvements in Word Error Rate (WER) on Librispeech and TEDLIUM-2.
These positive results indicate that incorporating semantic context can effectively improve transcription quality in streaming scenarios while keeping great performances using full context audio due the Dynamic Chunk Training.
Further research will aim to evaluate the proposed method with languages that have different linguistic structures and adapt the chunk size during inference depending on the linguistic and acoustic features of the input audio. Moreover, we aim to improve the training of the context module and the fine-tuning of the Sentence Embedding Model by using truncated text instead of the full audio transcription.
\section{Acknowledgements}
This work was also granted access to the HPC resources of IDRIS under the allocation 2025-A0191014876 made by GENCI.
\section{Ethical considerations and limitations}
We used Mistral-7B to generate paraphrases for fine-tuning the sentence-embedding teacher model. Because the pretraining corpora for Mistral-7B are not fully disclosed, we cannot exclude the possibility that public ASR corpora (e.g., LibriSpeech or TEDLIUM-2) overlap with its training data; this creates a theoretical risk of test-set contamination or memorized content being reintroduced via paraphrases. To mitigate this risk in our experiments, we (i) restrict the LLM use to an offline paraphrase-generation stage that is only used to fine-tune the teacher (the LLM and the teacher are never used at inference time) and (ii) limit the teacher’s fine-tuning to training-set transcripts only.
%
\section{Bibliographical References}\label{sec:reference}
\bibliographystyle{lrec2026-natbib}
\bibliography{lrec2026-example}

\section{Language Resource References}
\label{lr:ref}
\bibliographystylelanguageresource{lrec2026-natbib}
\bibliographylanguageresource{languageresource}

\end{document}